\documentclass[10pt,twocolumn,letterpaper]{article}

\usepackage{cvpr}
\usepackage{times}
\usepackage{epsfig}
\usepackage{graphicx}
\usepackage{amsmath}
\usepackage{amssymb}
\usepackage[utf8]{inputenc}
\usepackage{subfiles}
\usepackage{siunitx}
\usepackage{stfloats}
 
\usepackage[pagebackref=true,breaklinks=true,letterpaper=true,colorlinks,bookmarks=false]{hyperref}

\cvprfinalcopy 

\def\cvprPaperID{4626} 

\ifcvprfinal\fi
\begin{document}

\title{Inverse Graphics: Unsupervised Learning of 3D Shapes from Single Images}

\author{Talip Uçar \\
Department of Computer Science\\
University College London \\
{\tt\small ucabtuc@ucl.ac.uk}
}

\maketitle

\begin{abstract}
Using generative models for Inverse Graphics is an active area of research. However, most works focus on developing models for supervised and semi-supervised methods. In this paper, we study the problem of unsupervised learning of 3D geometry from single images. Our approach is to use a generative model that produces 2-D images as projections of a latent 3D voxel grid, which we train either as a variational auto-encoder or using adversarial methods. Our contributions are as follows: First, we show how to recover 3D shape and pose from general datasets such as MNIST, and MNIST Fashion in good quality. Second, we compare the shapes learned using adversarial and variational methods. Adversarial approach gives denser 3D shapes. Third, we explore the idea of modelling the pose of an object as uniform distribution to recover 3D shape from a single image. Our experiment with the CelebA dataset \cite{liu2015faceattributes} proves that we can recover complete 3D shape from a single image when the object is symmetric along one, or more axis whilst results obtained using ModelNet40 \cite{wu20153d} show the potential side-effects, in which the model learns 3D shapes such that it can render the same image from any viewpoint. Forth, we present a general end-to-end approach to learning 3D shapes from single images in a completely unsupervised fashion by modelling the factors of variation such as azimuth as independent latent variables. Our method makes no assumptions about the dataset, and can work with synthetic as well as real images. We present our results, by training the model using the $\mu$-VAE objective \cite{ucar2019bridging} and a dataset combining all images from MNIST, MNIST Fashion, CelebA and six categories of ModelNet40. The model is able to learn 3D shapes and the pose in qood quality and leverages information learned across all datasets. The model can be used for the classification of objects, generating new 3D shapes, recovering 3D shapes from given images, and rendering images of the same object from new viewpoints.

\end{abstract}

\section{Introduction}

Reconstruction of 3D structure from 2D images is an establised reasearch problem in computer vision. Traditionally, multiple views of an object are used to reconstruct its 3D representation \cite{furukawa2015multi,liu2010ray}. With recent progress in deep learning, there has been an interest in applying deep learning methods for inferring 3D shapes from 2D images \cite{yan2016perspective, choy20163d} and generating new 3D shapes from a learned representation \cite{gadelha20173d, zou20173d}. Taking inspiration from some of these works, this paper presents a framework to tackle the problems of inferring 3D geometry from a single 2D image (inverse graphics) using generative models. 

Although there have been numerous works on investigating generative models in the context of inverse graphics in recent years, they usually require assistance in the form of semi-, or full-supervision \cite{choy20163d, balashova2018structure, rezende2016unsupervised, girdhar2016learning, mandikal20183d, wu2016learning}. The supervision can be given implicitly (passive), and/or explicitly (active). Examples of active supervision include using 3D ground truth data (3D supervision), providing 2D images with annotation, and using multiple views of each object (2D supervision). Passive supervision is usually implicit in the form of restricting the dataset to contain images of objects in pre-determined poses, requiring images to be silhouettes,depth maps, synthetic, or any other form.

Most approaches require at least one of these supervision methods. The focus of this paper is to take the aforementioned methods one step further, to infer 3D shape from a single image without making any assumptions on training data while using both synthetic as well as real images (unsupervised in true sense). Thus, the mantra of this work is that \textit{a single image is all you need}.  

The paper is organized as following:First, we give a brief literature review. Then, in Section~\ref{model_and_training}, we present probabilistic 3D autoencoder, and details of training. Section~\ref{experiments} is dedicated to report the results of our experiments on learning 3D shapes from general datasets, the effects of modelling the azimuth as uniform random distribution and the results of the final model. We end the paper with a brief summary in Section~\ref{summary_and_discussion}.

\section{Related Work}

Recent years have seen a big interest in using deep learning techniques in areas of rendering (3D$\rightarrow$2D) and inverse-rendering (2D$\rightarrow$3D). The reconstruction of 3D shapes from the 2D input data and sampling new shapes, have been studied extensively in the context of generative models. Using autoencoders (AE) and variational autoencoders (VAE) to address these problems has been a common approach \cite{choy20163d, balashova2018structure,girdhar2016learning, mandikal20183d, nguyen2018rendernet, nash2017shape} while using generative adversarial networks (GANs) has also been proposed \cite{wu20153d, gadelha20173d}. But, they usually require some form of supervision such as: using 3D ground truth data \cite{rezende2016unsupervised, choy20163d, wu20153d, yan2016perspective}, utilizing multiple views of the object \cite{yan2016perspective}, using annotations\cite{girdhar2016learning, nguyen2018rendernet}, restricting the dataset to have silhouettes and/or depth maps of objects\cite{gadelha20173d, arsalan2017synthesizing, wiles2017silnet, yang2018learning}, or normal images\cite{nguyen2018rendernet}, and constraining the images to be rendered from particular view points\cite{gadelha20173d, wiles2017silnet, yang2018learning}. They are also mostly experimented on synthetic datasets. A short literature review is presented next. It is divided into two sections, based on the level of supervision used: i) Supervised, ii) Semi-supervised. 

\subsection{Supervised Methods:}

\begin{figure}[t]
\begin{center}
    \hbox{\includegraphics[width=1.0\linewidth]{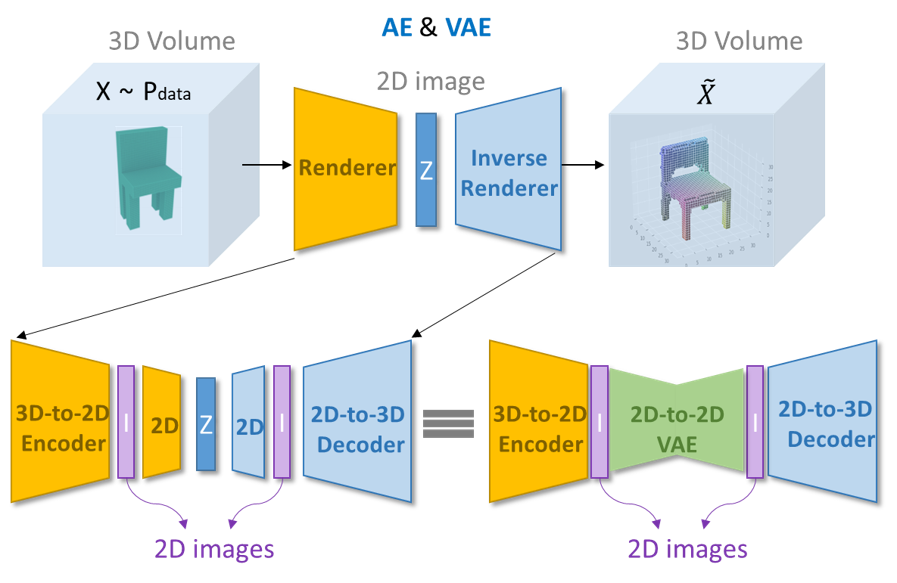}}

\end{center}
   \caption{Using autoencoders for rendering and inverse-rendering.}
\label{fig:vae_combinations}
\end{figure}
We can consider supervised methods under two sub-categories: 2D, and 3D supervision. Most work using full supervision has been done by using AEs and VAEs based on Convolutional and De-Convolutional Neural Networks (CNN, DeCNN respectively). The networks that work in 3D domain make use of volumetric convolutions (i.e. 3D-CNN for encoder, 3D-DeCNN for decoder). We can combine these networks in various ways to learn 2D$\rightarrow$3D, 3D$\rightarrow$2D, 3D$\rightarrow$3D, and 2D$\rightarrow$2D mappings as visualized in Figure~\ref{fig:vae_combinations}. The first three mappings require 3D ground truth data during training. The last one (2D$\rightarrow$2D) does not require 3D ground truth, and can be used for supervised, semi-supervised and unsupervised methods. For example, a 3D encoder can be paired with 3D decoder to learn 3D$\rightarrow$3D voxel mapping \cite{balashova2018structure}. We can replace 3D encoder with 2D one to have 2D$\rightarrow$3D mapping \cite{choy20163d}, or use 2D encoder in addition to 3D encoder to form a T-network and learn both 2D$\rightarrow$3D and 3D$\rightarrow$3D mappings \cite{girdhar2016learning, mandikal20183d}. Moreover, the bottleneck between encoder and decoder can be implemented as 3D-convolutional LSTM to feed single, or multiple images of same object from different viewpoints to improve results\cite{choy20163d}. We can also take advantage of key-points and annotations during training \cite{nguyen2018rendernet, nash2017shape}. Approaches using VAE might suffer from low quality of generated samples since VAEs learn data distribution indirectly by maximizing a lower-bound on log-likelihood rather than learning data distribution directly. Thus, taking a different approach, we can train 3D-DeCNN by using adversarial objective to implement a generative model (i.e. 3D-GAN) \cite{wu2016learning}.3D-GAN can be combined with a pre-trained VAE to infer latent vector from a 2D image \cite{wu2016learning, gadelha20173d}. This approach becomes a hybrid of 2D-VAE and 3D-GAN, in which each network is trained separately, to go from 2D image to 3D object. Training networks separately is sub-optimal. A better approach would be training whole architecture end-to-end. We can do this by using a GAN to map random samples from a latent $z$ to a 3D voxel grid, and then project it back to depth and silhouette images, and then to a final image \cite{zhu2018visual}. A big disadvantage of this approach is that it requires both 2D and 3D data. The approaches that do not utilize 3D ground truth data typically require a differentiable renderer (or, projection unit) to project inferred 3D representation onto an image plane \cite{wiles2017silnet,gadelha20173d, yang2018learning, yan2016perspective}. They might also require extra data in the form of silhouettes \cite{wiles2017silnet,gadelha20173d, gwak2017weakly, kanazawa2018learning}, or depth maps of images, and/or pose, lighting, or key-point annotations \cite{wiles2017silnet,yang2018learning,kanazawa2018learning}. The projection can be done by deciding whether a volume is occupied along the direction of rays to the pixel plane \cite{wiles2017silnet,gadelha20173d}. As a separate approach, 3D shapes can be inferred by taking advantage of multi-view observations of the object in the form of foreground masks, depth, or color images \cite{tulsiani2017multi, tulsiani2018factoring, arsalan2017synthesizing, gwak2017weakly, yan2016perspective}. Similarly, the shapes can be reconstructed by using images from same, or different categories, annotated with camera poses \cite{yang2018learning}.

\subsection{Semi-supervised Methods:}
A variational framework proposed by \cite{rezende2016unsupervised} is used to train a generative model of 3D objects represented as volumes, or meshes. The 3D ground-truth data is used for training when it is available, and otherwise, they require a differentiable, or off-the-shelf renderer (openGL). Their experiments don't go beyond using simple shapes such as spheres and cubes. Also, using mixture of methods for rendering is a major drawback. A good model should be able work with any dataset without changing any parts of the model due to nature of the dataset.

\begin{figure*}
\begin{center}
\includegraphics[width=1.0\linewidth]{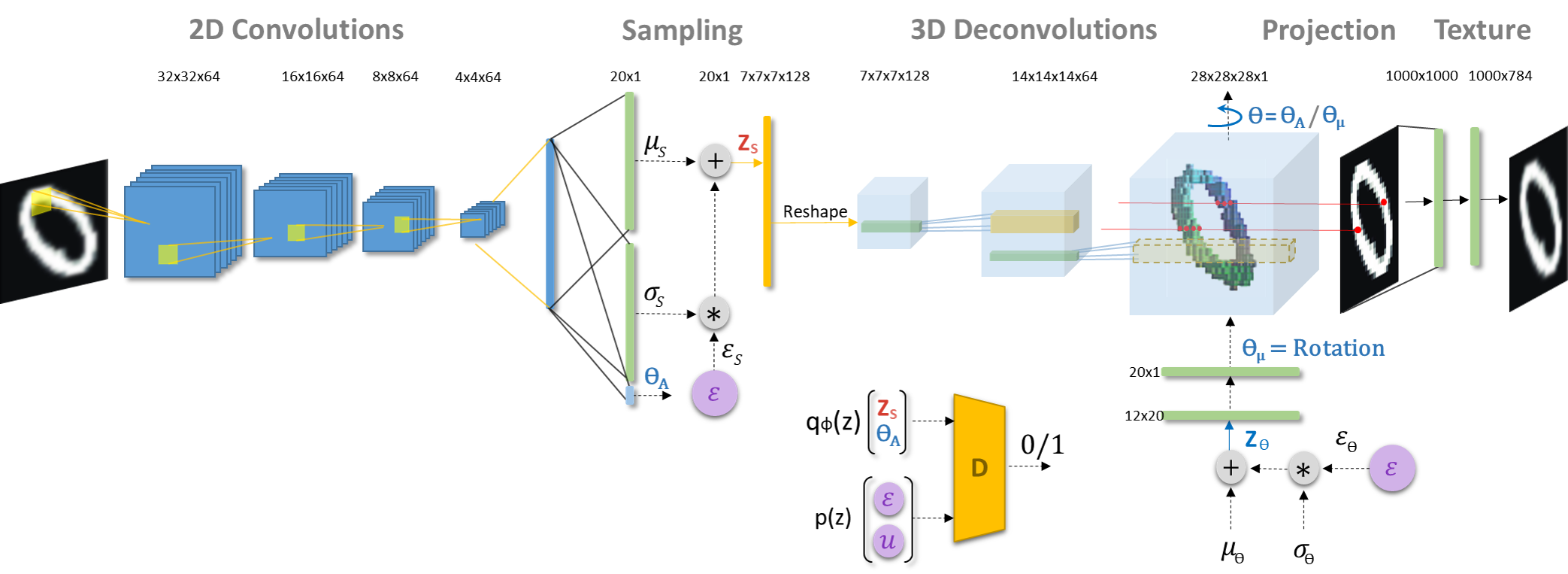}
\end{center}
   \caption{Probabilistic 3D-Autoencoder Model.}
\label{fig:final_model}
\end{figure*}

\section{Probabilistic 3D-Autoencoder Model}\label{model_and_training}

We developed a probabilistic 3D-Autoencoder to be trained with either variational\cite{kingma2014adam}, or adversarial methods\cite{makhzani2015adversarial}. In the variational approach, we train a 3D-AE with three different objective functions to make comparisons: i) Standard VAE objective with KL divergence (VAE) \cite{kingma2014adam}, ii) $\beta$-VAE \cite{higgins2017beta}, and iii) $\mu$-VAE objective \cite{ucar2019bridging}. The adversarial approach is implemented by using an additional discriminator to regularize the latent layer (AAE) \cite{makhzani2015adversarial}. 

Both variational and adversarial methods include a reconstruction loss as part of their objective. The difference comes in the form of how each regularizes the parameters of latent variables. For example, $\beta$-VAE penalizes KL term by multiplying it with a coefficient $\beta>1$. $\mu$-VAE keeps an aggregate mean of latent samples same as that of prior (i.e. zero mean) while allowing samples to spread out, mitigating posterior collapse commonly observed in VAEs with high-capacity decoders \cite{ucar2019bridging, razavi2019preventing, chen2016variational, dieng2018avoiding, kim2018semi,van2017neural,bowman2015generating, kingma2016improved, sonderby2016train, zhao2017towards}. Adversarial approach uses a discriminator to regularize the latent space by comparing samples from $q_\phi(z)$ to the ones from a prior distribution $p(z)$.

The base model has three parts: i) A 2D convolutional encoder that maps images to the parameters of the latent variable $Z_S$ used for shape, ii) A 3D de-convolutional decoder that maps samples drawn from latent layer to 3D voxel grid and iii) A differentiable renderer proposed by \cite{gadelha20173d} to render images from 3D voxel grid. Moreover, failing to capture local structures is a known problem in latent variable models \cite{larsen2015autoencoding, razavi2019preventing}. To mitigate this problem, an optional two layer fully connected neural network, referred as a texturizer, is added to the output of projection layer to improve learning shades and texture in images. When, the texturizer is used, the output of projection layer learns the shape of the object while the texturizer learns shades and textures in the image. Finally, for adversarial training (AAE), a discriminator is added to the base model. The model used for variational and adversarial training are shown in Figure~\ref{fig:final_model}. It is trained end-to-end to minimize the loss functions shown in Table~\ref{table:objective_for_3D} to compare different approaches. 

The final 3D-VAE proposed in this work uses the base model with texturizer. Additionally, it trains one more latent variable for the azimuth, $\theta_{\mu}$, independent of the shape, $Z_S$. During training, we would ideally maximize the marginal likelihood of the data:

\begin{equation}
    \resizebox{0.5\textwidth}{!}{$P_{\theta}(x_{2D}) = \iint P(x_{2D} \vert x_{3D}, \theta_{\mu})P_{\theta}(x_{3D} \vert z_s)P_\lambda(\theta_{\mu} \vert z_{\theta})P(z_{\theta})P(z_s) \,dz_s\,dz_{\theta}$}
\end{equation}

\noindent where $\theta$ in $P_\theta$ refers to the parameters of the decoder while it is used to refer to the azimuth in the rest of the equation. $x_{2D}$, $x_{3D}$, and $\lambda$ correspond to 2D image, 3D voxel, and the parameters of the generative model used for the azimuth respectively. Since this formulation is intractable, we take advantage of commonly used variational  methods \cite{kingma2013auto, rezende2014stochastic, hoffman2013stochastic} although it should be noted that $\mu$-VAE replaces the KL term used in evidence lower bound (ELBO) objective \cite{kingma2013auto}.We should further note that while the parameters $\mu_s$ and $\sigma_s$ are modelled as a function of data (amortized stochastic variational inference (SVI)), the parameters $\mu_{\theta}$ and $\sigma_{\theta}$ are not. We could have used the same encoder for the latter parameters, but we wanted to avoid any entangling between the shape and the azimuth through parameters of the encoder. We could also use a separate encoder for amortized training of $Z_{\theta}$, but the pose should not ideally be a function of data. The final model is trained using $\mu$-VAE objective function since $\mu$-VAE allows to spread out samples in a controlled fashion, which opens the door to train images from multiple datasets at the same time while avoiding the posterior collapse. 

\begin{table*}[ht]
\centering
\caption{Objective functions. The $\mu$-VAE includes $\mu_{\theta}$ and $\sigma_{\theta}$ parameters for the case when the azimuth, $\theta$ is modelled with a latent model.}  
    \hspace*{0cm}\begin{tabular}[t]{ll}
        \hline
        &\multicolumn{1}{c}{\textbf{Objective}}\\
        \hline
        \hline
        
        \textbf{VAE}&$\mathcal{L}_{e} = \mathbb{E}_{q_\phi(z \vert x)} \left[\log p_\theta(x \vert z) \right] -  \mathrm{KL}(q_\phi(z\vert x) \Vert p(z))$  \\
        
        \textbf{$\beta$-VAE}&$\mathcal{L}_{\beta} = \mathbb{E}_{q_\phi(z \vert x)} \left[\log p_\theta(x \vert z) \right] -  \beta * \mathrm{KL}(q_\phi(z\vert x) \Vert p(z))$ \\
        
        \textbf{AAE}&$\mathcal{L}_{\mathrm{a}} = \mathbb{E}_{q_\phi(z \vert x)} \left[\log p_\theta(x \vert z) \right] - \lambda*\mathrm{KL}(q_\phi(z) \Vert p(z))$ \\
        
        \textbf{$\mu$-VAE}&$\mathcal{L}_\mathrm{\mu}=\mathbb{E}_{q_\phi(z \vert x)} \left[\log p_\theta(x \vert z_s,z_{\theta}) \right] - \frac{1}{B} 
        \left[ 
        \lvert\sum_{i=1}^B\sum_{d=1}^D\mu_{S_d}^{(i)}\rvert  + 
        \lvert\sum_{i=1}^{B}\sum_{d=1}^{D}\mu_{\theta_d}^{(i)}\rvert  +
        \sum_{i=1}^{B}\sum_{d=1}^{D}\left[\log\sigma_{S}^2\right]_d^{(i)} + \left[\log\sigma_{\theta}^2\right]_d^{(i)}  
        \right]$
           \\
        \hline

    \end{tabular}
    \label{table:objective_for_3D}
\end{table*}%




\subsection{The 2D-Encoder \& 3D-Decoder}
A 2D-Encoder, based on CNNs, is used to map 2D images to the mean, $\mu$, and the standard deviation, $\sigma$, parameters of a multivariate Gaussian. The decoder is based on 3D-DeCNN, inspired by \cite{wu2016learning}. The input to the decoder is the samples from latent variable $Z_S$. Decoder transforms $Z_S$ into a $28^3$ volume to represent 3D shape in the form of voxels ($\textbf{v}$). 3D shape has a binary representation, where each voxel $v \in [0, 1]$ represents whether it is occupied. 2D CNN layers use 3x3 kernels with strides of 2 while 3D DeCNN layers utilizes 3x3x3 3D-kernels with strides of 2x2x2. Batch normalization (batchnorm) \cite{ioffe2015batch} and Leaky ReLu are used for all layers, except the final layers of the encoder (linear), and the decoder (sigmoid).

\subsection{Projection:}

\begin{figure}[!ht]
    \begin{center}
    \includegraphics[keepaspectratio, width=0.2\paperwidth]{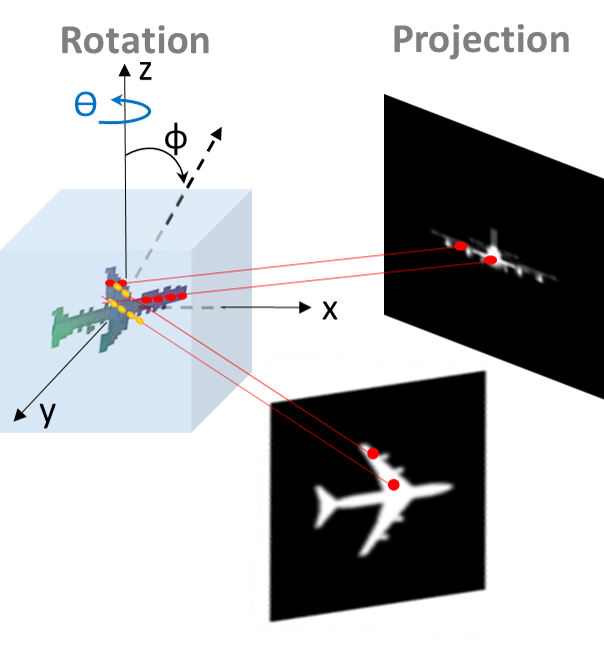}
   \caption{\textbf{Projection:} Defining parameters of projection.}
   \label{fig:projection_params}
    \end{center}
\end{figure}

In this work, the objects are assumed to be at the origin, upright oriented, and the scale is fixed. Only variation is assumed to be on the azimuth plane.We ran experiments modelling the azimuth as a constant, or variable: 

\begin{enumerate}
    \itemsep0em 
    \item \textbf{Azimuth as a constant:} It is considered zero for both variational and adversarial training.
    \item \textbf{Azimuth as a variable:} It is modelled as an independent variable in two ways:
   \begin{enumerate}
        \itemsep0em 
        \item \textbf{Random uniform noise:} In some experiments, we modelled it as an random uniform noise variable sampled from range $[-\pi,\pi]$ to induce learning 3D geometry. In this case, the azimuth is not learned, and is used to enforce rendering of the same image from any viewpoint. This method is useful to recover 3D shapes of the symmetric objects such as human head by using only the face images as in CelebA dataset.
     
         \item \textbf{Independent latent variable:} In the final model developed in this paper, the azimuth is modelled with a second generative model. We use an independent latent variable with parameters $\mu_{\theta}$ and $\sigma_{\theta}$ as well as two-layer neural network with \textbf{tanh} activation. Output ($[-1,1]$) of the final layer is scaled to $[-\pi,\pi]$ and used to change viewpoint directly (i.e. $\theta_{\mu}$ is a continuous variable).
    \end{enumerate}
\end{enumerate}

The projection layer is used to render images from 3D voxel grid, $V$. If the azimuth is modelled as a variable, then the projection layer changes viewpoint by using a rotation matrix with parameters ($\theta, \phi=0$), resulting in new viewpoint $V_{\theta, \phi}$. This operation uses nearest neighbor sampling through the floor operator. Once the viewpoint is changed, the voxels are projected onto the image plane to render an image as shown in Figure~\ref{fig:projection_params}. For projection, a simple differentiable projection operator proposed by \cite{gadelha20173d} is used:
\begin{equation}
    P_{\theta, \phi}((i,j),V) = 1 - \exp^{-\sum_{k} V_{\theta, \phi}(i,j,k)}
\end{equation}

This formulation uses an exponential function to have a smooth and differentiable function, where the exponent is sum over voxel values along the direction of rays from corresponding pixels. Projected value is zero when there is no voxel occupied while it approaches 1 as the number of occupied voxels increases.

\subsection{Dataset}
Four datasets are used throughout this work: MNIST, MNIST Fashion, CelebA, and ModelNet40 \cite{wu20153d}. CelebA has around 200k images, from which we selected 10k quality images. They are converted to gray scale, cropped and resized to 28x28 for faster experimentation although it becomes a much harder task to learn geometry due to limited resolution in both image, and voxel space. ModelNet40 dataset has objects from 40 categories. We chosen three convex-shaped objects (airplane, car and person) and three concave-shaped ones (cup, bowl, and chair). Between the two categories, the convex shapes are easier to learn 3D shapes from. In this dataset, the rendered images are taken from 12 evenly spaced viewing angles with orthographic projection. Hence, the azimuth is divided into 12 bins i.e. $\theta=[0,30, ...,330]$. The images are scaled down to 28x28. It should be emphasized that although each object in particular category of ModelNet40 has views from 12 different angles, the models are trained with a single image selected from a randomly shuffled dataset, and do not depend on the number of view points.


\subsection{Training}
In all four objective functions, a spherical 2-D Gaussian prior distribution is imposed on the hidden codes $z$. For AAE, the encoder $q_\phi(z|x)$ acts as the generator and tries to fool the discriminator by generating samples from the aggregated posterior distribution $q(z)$ that are similar to those from the prior distribution p(z). Hence, the adversarial training encourages $q(z)$ to match to $p(z)$. For the $\beta$-VAE, we used $\beta=30$. 

In the final model,  since the azimuth, $\theta$, is modelled as an output of a second generative model, we used a separate optimizer to learn $\mu_\theta$ and $\sigma_\theta$ parameters of the latent variable, $Z_\theta$ directly by back-propagating errors while freezing VAE parameters. One might choose to update $\mu_\theta$ and $\sigma_\theta$ parameters multiple times per each encoder-decoder update since their training is not amortized. But, in this work, all parameters are updated once per iteration. 

\textbf{Optimization:} Adam algorithm with high momentum ($\beta1 = 0.9, \beta2=0.999$) is used as the optimizer throughout this work. High momentum is chosen mainly to let most of previous training samples influence the current update step. For reconstruction loss, mean square error, $\Vert x-x' \Vert^2$, is used for all cases. If the texturizer is used, then reconstruction loss has two terms, one from the output of the projection, and another from the output of the texturizer. In each term, the error is computed using the same input image X:

\begin{equation}
    L_{recon} = \Vert X-X'_{projection} \Vert^2 + \Vert X-X'_{texture} \Vert^2\\
\end{equation}

\section{Experiments}\label{experiments}
\subsection{Learning 3D shapes from general datasets}\label{experiments_general_dataset}
\begin{figure}[t]
\begin{center}
   \includegraphics[width=1.0\linewidth]{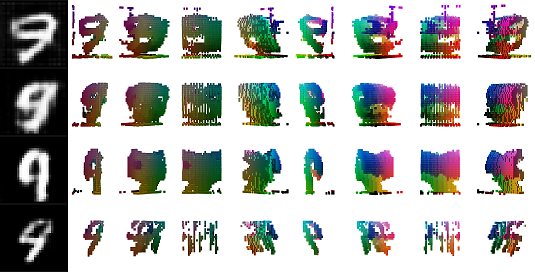}
\end{center}
\caption{\textbf{Comparing geometries using digit 9 from MNIST:} From top to bottom: VAE, $\beta$-VAE, AAE, $\mu$-VAE.}
\label{fig:comparing_geometry}
\end{figure}

\begin{figure}[t]
\begin{center}
   \includegraphics[width=1.0\linewidth]{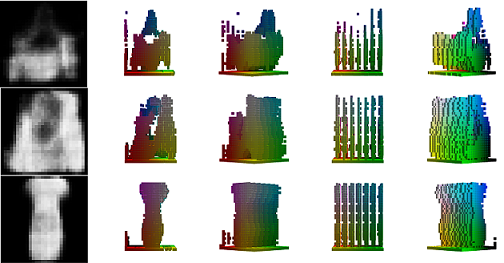}
\end{center}
   \caption{\textbf{Random samples:} Random samples from the AAE model trained with MNIST Fashion. From top to bottom, the shapes are: bag, dress, dress.}
\label{fig:random_samples_mnist_mfashion}
\end{figure}

We explored the idea of learning representations of 3D shapes from general datasets such as MNIST and MNIST Fashion. The model is separately trained on each dataset by using four objective functions. The azimuth is assumed fixed (i.e. zero), and the texturizer is used only for MNIST Fashion. The adversarial objective learns a dense representation while $\mu$-VAE encourages sparsity as shown in Figure~\ref{fig:comparing_geometry}. Since there is no views along the depth dimension, the shapes are learned as slices of layers, in which the shadows and texture are represented by later layers along the depth (see top row, the handle of the bag, in Figure~\ref{fig:random_samples_mnist_mfashion}).

\subsection{Modelling $\theta$ with a uniform distribution:}\label{azimuth_noise}
\begin{figure}[t]
\begin{center}
    \hbox{\includegraphics[width=1.0\linewidth]{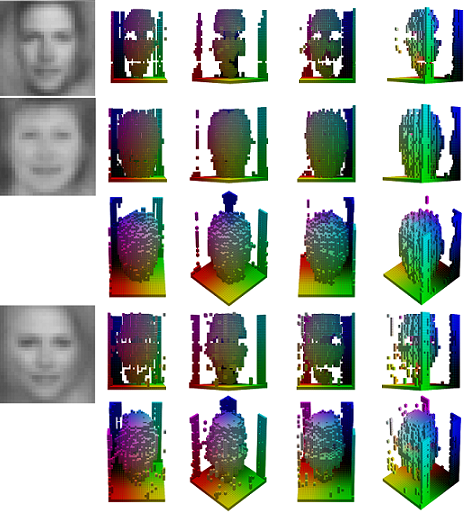}}

\end{center}
   \caption{\textbf{AAE with texturizer:} Closer look at learned 3D structure of sampled faces from various azimuth angles.}
\label{fig:aae_3d_closer_look}
\end{figure}

\begin{figure}[t]
\begin{center}
   \includegraphics[width=1.0\linewidth]{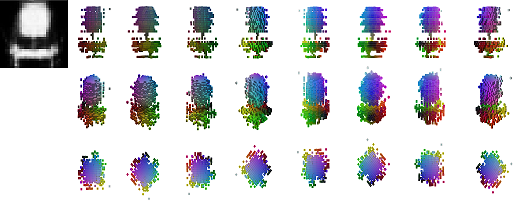}
\end{center}
   \caption{\textbf{The side-effect of modelling the azimuth as uniform random noise ($\theta$=random uniform [$-\pi,\pi$]):} From top to bottom, the chair is shown at elevation: 0, 30 and 90 degrees.}
\label{fig:efect_of_noise2}
\end{figure}

When the azimuth is fixed as zero, the model learns the shapes along the depth dimension in sliced layers. This is because we don't have access to images taken from different viewpoints for datasets such as MNIST, MNIST Fashion and CelebA. Noting that most objects are symmetrical along one of the axis, we explored the idea of modelling the pose as uniform random distribution around the axis of symmetry. We trained the AAE model with texturizer on CelebA dataset and modelled the azimuth as a random uniform distribution. Note that , in AAE training, the azimuth is one of the outputs of encoder, and it is regularized by using uniform distribution prior ($[-1,1]$) and a discriminator (see Figure~\ref{fig:final_model}). CelebA includes only face images with little variations in the azimuth, but the shape of head is symmetric along the z-axis. We were able to recover 3D shape of a human head as seen in Figure-\ref{fig:aae_3d_closer_look}. In some cases, the model was able to learn local features such as the eyes and the mouth. However, this approach may not always work as intended. For example, we trained the VAE model on chairs category of ModelNet40. This time, the azimuth is modelled as a random uniform noise. As shown in Figure~\ref{fig:efect_of_noise2}, the shape of the chair is learned in a way that the model would be able to render the same image from any viewpoint. In other words, the shape learned is circular around z-axis.

\subsection{Learning convex and concave shapes:}\label{learning_concave_shapes}
\begin{figure}[t]
\begin{center}
    \hbox{\includegraphics[width=1.0\linewidth]{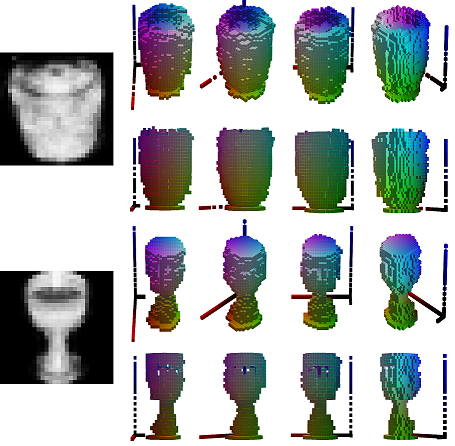}}

\end{center}
   \caption{\textbf{AAE with texturizer:} Random samples of learned 3D structure from Cup category of ModelNet40.}
\label{fig:aae_random_samples_cup_modelnet40}
\end{figure}

\begin{figure}[t]
\begin{center}
    \hbox{\includegraphics[width=1.0\linewidth]{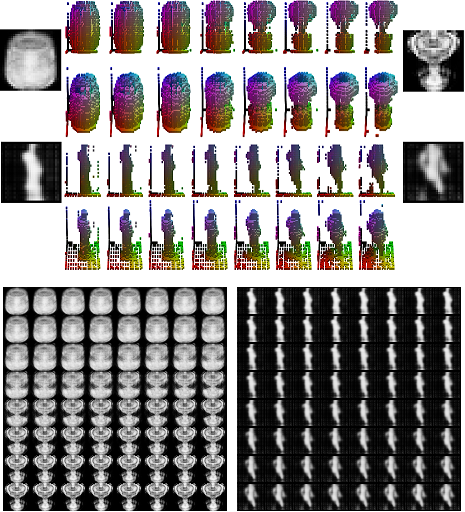}}

\end{center}
   \caption{\textbf{AAE with texturizer:} Latent traversal in 3D (top) as well as corresponding renderings (bottom) for cup and person categories of ModelNet40.}
\label{fig:aae_interpolations_modelnet40}
\end{figure}

Concave shapes such as cups are not easy to learn, especially with the voxel representation. As to test the limits of voxel representation, the AAE model with texturizer is trained on the cups and persons category of ModelNet40. The azimuth is regularized with uniform distribution using discriminator as before. Figure~\ref{fig:aae_random_samples_cup_modelnet40} shows two random samples from the cups category. The model is able to understand the depth of the cup in the first sample while it fails in the second one. Moreover, the latent space is smooth as seen in latent traverses in Figure~\ref{fig:aae_interpolations_modelnet40}.

\subsection{The final model -- $\mu$-VAE with $Z_\theta$:}\label{final_model_uvae}
\begin{figure}[t]
\begin{center}
    \hbox{\includegraphics[width=1.0\linewidth]{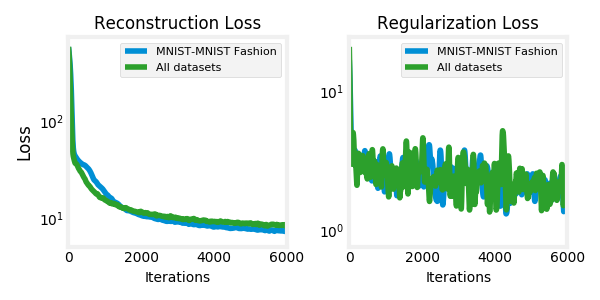}}
\end{center}
   \caption{\textbf{Final $\mu$-VAE model:} The reconstruction and regularization loss while training the combined dataset of all datasets (MNIST, MNIST Fashion, CelebA and ModelNet40) and two (MNIST and MNIST Fashion).}
\label{fig:loss_curves}
\end{figure}

\begin{figure}[t]
\begin{center}

    \hbox{
    \includegraphics[width=0.33\linewidth]{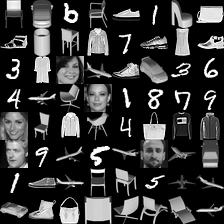}
    \includegraphics[width=0.33\linewidth]{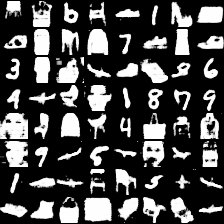}
    \includegraphics[width=0.33\linewidth]{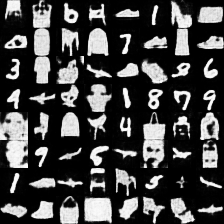}
    }
    
    \hbox{
    \includegraphics[width=0.33\linewidth]{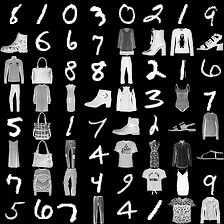}
    \includegraphics[width=0.33\linewidth]{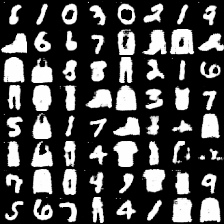}
    \includegraphics[width=0.33\linewidth]{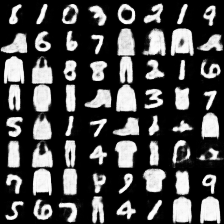}
    }
    
\end{center}
   \caption{Input data, and reconstructions at the projection and texturizer. 
   \textbf{Top:} The model is trained on combined dataset of MNIST, MNIST Fashion, CelebA and ModelNet40. \textbf{Bottom:} The model is trained on combination of MNIST and MNIST Fashion.} 
   \label{fig:uvae_reconstruction_all_datasets}
\end{figure}

\begin{figure*}
\begin{center}
\includegraphics[width=1.0\linewidth]{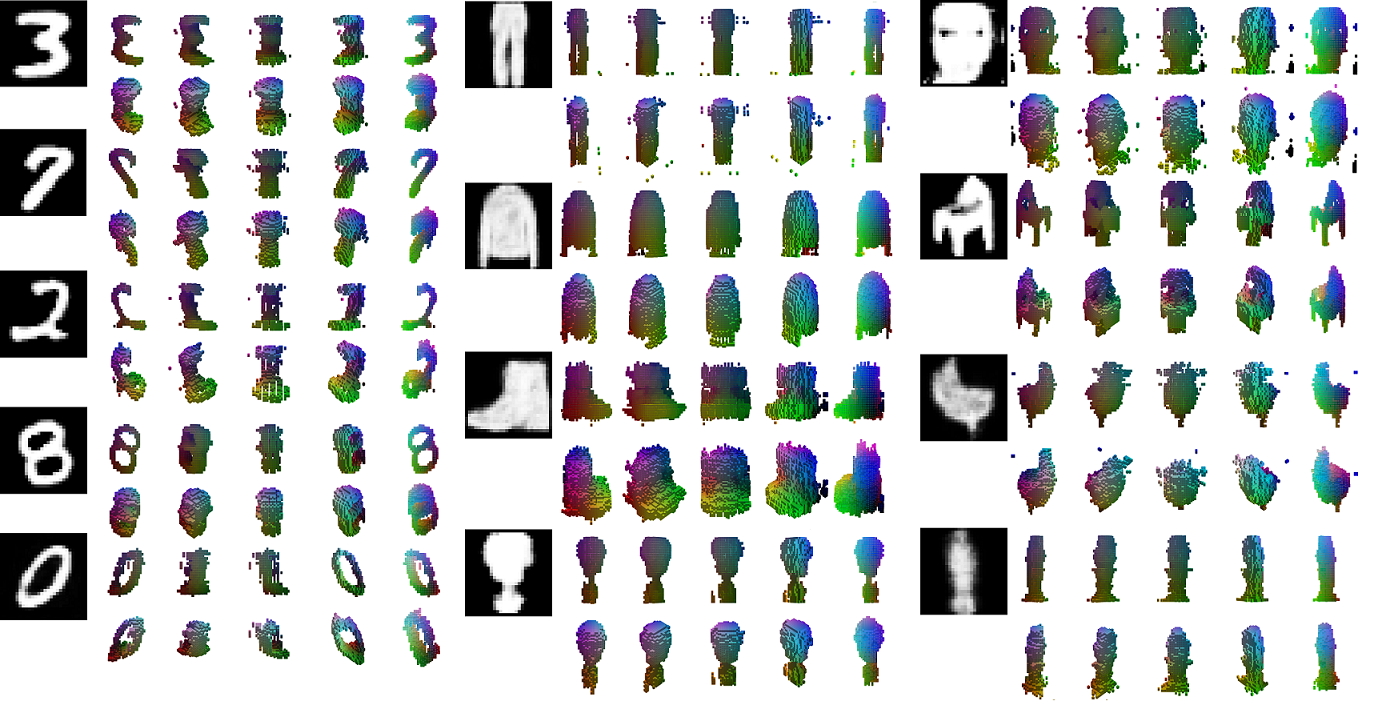}
\end{center}
   \caption{\textbf{Final $\mu$-VAE model:} 3D shapes shown at $\ang{0}$ and $\ang{30}$ elevations from different azimuth angles. Corresponding image is rendered using both $\theta,\phi=0$.  From top to bottom, the objects are;
   \textbf{Left column:} Digits 3, 7, 2, 8, 0 from MNIST,
   \textbf{Middle column:} Pants, shirt, and shoe from MNIST Fashion, cup from ModelNet40,
  \textbf{ Right column:} Human head from CelebA; chair, chair, and person from ModelNet40. A demo of how the shapes are learned during training can be seen at: https://pilatracu.github.io/3dvae/}
\label{fig:3D_using_all}
\end{figure*}

\begin{figure}[htb]
\begin{center}
    \hbox{
    \includegraphics[width=0.33\linewidth]{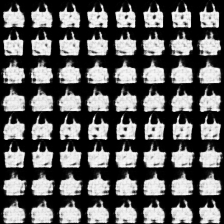}
    \includegraphics[width=0.33\linewidth]{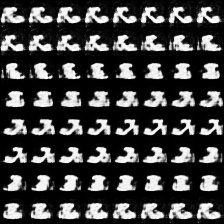}
    \includegraphics[width=0.33\linewidth]{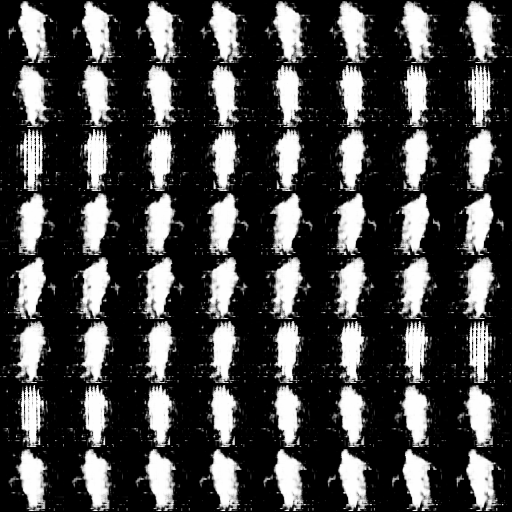}

    }
    \hbox{
    \includegraphics[width=0.33\linewidth]{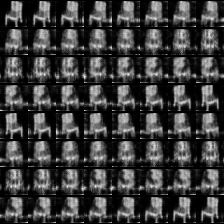}
    \includegraphics[width=0.33\linewidth]{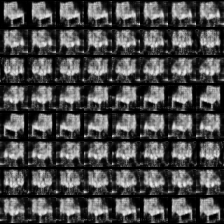}
    \includegraphics[width=0.33\linewidth]{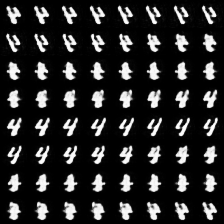}
    }
\hbox{
    \includegraphics[width=0.33\linewidth]{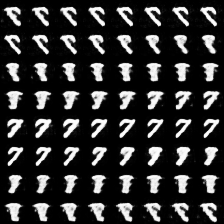}
    \includegraphics[width=0.33\linewidth]{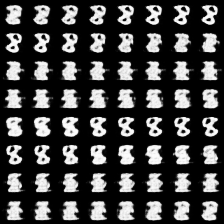}
    \includegraphics[width=0.33\linewidth]{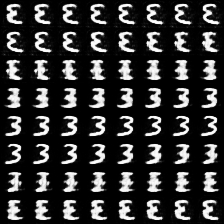}
    }
\end{center}\caption{\textbf{Final $\mu$-VAE model:} Rendering images from different viewpoints by sweeping $\theta$ across $[-\pi,\pi]$ in 64 steps (the top-left: $-\pi$, the bottom-right: $\pi$ in each image). \textbf{Top:} dress, shoe, person; \textbf{Middle:} chair, chair, digit 4; \textbf{Bottom:} digits 7, 8, 3.}
\label{fig:uvae_mn_mnfashion_rotation}
\end{figure}

\begin{figure}[htb]
 \begin{center}
     \hbox{\includegraphics[width=1.0\linewidth]{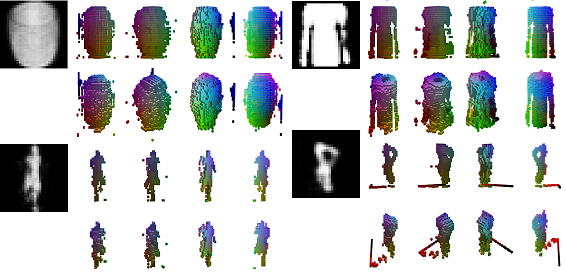}}
 \end{center}
  \caption{\textbf{Final $\mu$-VAE model:} Random samples shown at elevation=0, and 30. The top row: cup,  shirt; the bottom row: running man, digit 8.}
 \label{fig:uvae_random_samples}
 \end{figure}

The one advantage of $\mu$-VAE is that it allows us to spread out samples in latent space in a controlled way so that we can train multiple datasets using a single model at one-shot, potentially enabling the model to leverage what it learns from one dataset to learn better representations in others (i.e. transfer learning).Moreover, a true unsupervised model should be able to learn distributions of factors of variation such as azimuth, elevation, and lighting. As a proof of concept, we modelled the azimuth using a second generative model, in which the latent variable is modelled as multivariate Gaussian with diagonal co-variance (see Figure~\ref{fig:final_model}). The parameters $\mu_{\theta}$ and  $\sigma_{\theta}$ are learned directly using SVI while the $\mu_{S}$ and  $\sigma_{S}$ parameters of the shape are trained using amortized SVI. The model is trained using images from all datasets at one-shot. Figure~\ref{fig:loss_curves} shows the loss curves during training. Comparing renderings of the 3D shapes at outputs of projection layer and texturizer in Figure~\ref{fig:uvae_reconstruction_all_datasets}, we see that the model is able to render shapes of objects from all datasets in good quality despite of the fact that these datasets are very different. Moreover, the model learns 3D shapes of all objects in the way how humans would visualize as shown Figure~\ref{fig:3D_using_all}. The results are surprising especially for cases, in which we have only a single viewpoint of objects (for example, in MNIST, MNIST Fashion and CelebA). The model is able to learn true shapes and the azimuth for digits and fashion items using only front view of the objects (Figure~\ref{fig:3D_using_all}). We are able to render the digits and clothing from different viewpoints as seen in Figure~\ref{fig:uvae_mn_mnfashion_rotation}. Moreover, we trained the model using mixed images from MNIST and MNIST Fashion in a separate experiment to see whether the model could still be able to generate new viewpoints. We confirmed that it can. Thus, modelling the azimuth with a separate latent variable conditions the model such that we can render new images from new viewpoints regardless of whether the dataset contains images from multiple views. Finally, the trained model can be used to generate new quality shapes (Figure~\ref{fig:uvae_random_samples}).

\section{Discussion}\label{summary_and_discussion}
\begin{figure}[htb]
\begin{center}
    \hbox{\includegraphics[width=1\linewidth]{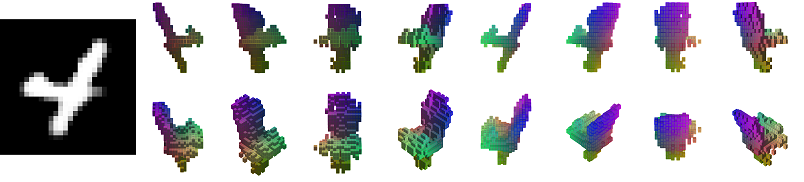}}
\end{center}
   \caption{\textbf{Final $\mu$-VAE model:} A case showing that inferring 3D shape from a 2D image could be tricky. In this case, the possible 3D objects are a plane, a chair, or digit 4.}
\label{fig:uvae_interesting}
\end{figure}

In this work, we investigated unsupervised learning of 3D shapes from single images, using diverse set of datasets. We noted the difference in 3D shapes learned by variational and adversarial methods, and suggested  a way to take advantage of symmetry of object to recover their shapes. We further developed a 3D generative model that can learn 3D shape and pose representations from single images using any given dataset. We also explored the idea of learning from a diverse set of datasets at one-shot. This enables the model to learn rich representations. For example, Figure~\ref{fig:uvae_interesting} poses an interesting case, in which the rendered image could be a projection of a digit 4, a chair, or a plane. However, when we observe the corresponding shape from $\ang{30}$ elevation, we see that the model generates shape in the form of a chair, one of the possibilities, although the shape could have been any of those three categories when observed from $\ang{0}$ elevation. This is very similar to how humans would visualize this image. If we trained the model using only MNIST, the shape generated would perhaps resemble digit 4 from any angle. Thus, training generative models using many datasets can bring us closer to bridging the gap between how humans and these models think about shapes. This would also enable us to use the latent representations in tasks such as classification. 


\clearpage
\clearpage


{\small
\bibliographystyle{ieee}
\bibliography{egbib}
}

\end{document}